\documentclass{article} % For LaTeX2e
\usepackage[final]{colm2026_conference}

\usepackage{amsmath}
\usepackage{amsmath}
\usepackage{microtype}
\usepackage{hyperref}
\usepackage{url}
\usepackage{booktabs}
\usepackage{graphicx}
\usepackage{subcaption}
\usepackage{amstext}
\usepackage{lineno}

\definecolor{darkblue}{rgb}{0, 0, 0.5}
\hypersetup{colorlinks=true, citecolor=darkblue, linkcolor=darkblue, urlcolor=darkblue}

\title{Beyond Accuracy: Diagnosing Algebraic Reasoning Failures in LLMs Across Nine Complexity Dimensions}

\author{Parth Patil, Dhruv Kumar \& Yash Sinha  \\
Department of Computer Science \& Information Systems\\
BITS Pilani \\
\texttt{\{h20201854, dhruv.kumar,yash.sinha\}@pilani.bits-pilani.ac.in} \\
\And
Murari Mandal\\
School of Computer Engineering,  \\
KIIT Bhubaneshwar \\
\texttt{\{murari.mandalfcs\}@kiit.ac.in} \\
\AND
}

\begin{document}

\ifcolmsubmission
\linenumbers
\fi

\maketitle

%%, ------------------------------------------------------------
\begin{abstract}

Algebraic reasoning remains one of the most informative stress tests for large language models, yet current benchmarks provide no mechanism for attributing failure to a specific cause. When a model fails an algebraic problem, a single accuracy score cannot reveal whether the expression was too deeply nested, the operator too uncommon, the intermediate state count too high, or the dependency chain too long. Prior work has studied individual failure modes in isolation, but no framework has varied each complexity factor independently under strict experimental control. No prior system has offered automatic generation and verification of problems of increasing complexity to track model progress over time. We introduce a nine-dimension algebraic complexity framework in which each factor is varied independently while all others are held fixed, with problem generation and verification handled by a parametric pipeline requiring no human annotation. Each dimension is grounded in a documented LLM failure mode and captures a structurally distinct aspect of algebraic difficulty, including expression nesting depth, simultaneous intermediate result count, sub-expression complexity, operator hardness, and dependent reasoning chain length. We evaluated seven instruction-tuned models spanning 8B to 235B parameters across all nine dimensions and find that working memory is the dominant scale-invariant bottleneck. Every model collapses between 20 and 30 parallel branches regardless of parameter count, pointing to a hard architectural constraint rather than a solvable capacity limitation. Our analysis further identifies a minimal yet diagnostically sufficient subset of five dimensions that together span the full space of documented algebraic failure modes, providing a complete complexity profile of a model's algebraic reasoning capacity.
\end{abstract}

%%, ------------------------------------------------------------
\section{Introduction}

Mathematical reasoning has become a central test bed for LLMs, with benchmarks
spanning grade-school arithmetic~\citep{cobbe2021gsm8k}, competition
mathematics~\citep{hendrycks2021measuring}, graduate-level applied
problems~\citep{fan2024hardmath}, and symbolic
computation~\citep{lample2020deep,saxton2019analysing}. \citet{song2026}
synthesise over 200 papers into a taxonomy of LLM reasoning failures. Despite
this breadth, these benchmarks share a fundamental limitation: they collapse all
sources of difficulty into a single accuracy score. Expression depth, operator
type, branching structure, and counting demand all co-vary across problems. And when
a model fails, the score records that it failed, but not whyit failed.

This matters because the failure modes are mechanistically distinct and have been
studied only in isolation. \citet{dziri2024faith} traced the steep drop from
59\% on three-digit multiplication to 4\% on four-digit to $O(n^2)$ cross-token
interactions. Also the counting failures originate in tokenisation and positional encoding
limits, not arithmetic gaps~\citep{zhang2024tokenization,chang2024}.
Compositional breakdown occurs specifically when sub-problems are chained, which are also not
solved in isolation~\citep{zhao2024}. No prior work has systematically varied each factor in isolation under strict experimental control. Existing studies do not hold the remaining dimensions fixed during such comparisons. Additionally, the model sets used in prior work are not large enough to reliably distinguish universal architectural limits from model-specific behaviors.

A second challenge is longevity. Static benchmarks saturate as models improve. They
lose its diagnostic value precisely when they are most needed.What is needed is a generation system that can automatically produce verified problems at any desired complexity level. This system must be capable of generating such problems on demand. Increasing the difficulty ceiling should require only a single parameter change. It should not require rebuilding the entire problem corpus from scratch. This paper addresses both challenges. We
define nine dimensions of algebraic complexity: syntactic length, tree depth,
operator hardness, working memory load, compositional branching, solution
ambiguity, counting load, sequential chain length, and numeric magnitude; each
grounded in a documented failure mode with prior empirical support. We then build
a parametric pipeline that varies each dimension independently while holding the
other eight fixed. And every problem is CAS-verified before evaluation.

We evaluated seven instruction-tuned models spanning 8B to 235B parameters across
all nine dimensions. Working memory (D4) is a scale-invariant architectural
limit: every model collapses between 20 and 30 parallel branches regardless of
parameter count. Sequential chaining (D8), tested to 12 steps, is far more
destructive than prior two-model studies indicated. Counting load (D7) reveals
the widest model divergence: Claude 3.5 Haiku holds 100\% at $K{=}300$ while
Llama 3 8B fails from $K{=}25$ in counting load. Five dimensions (D2, D4, D5, D7, D8) are jointly
sufficient to characterise a model's full algebraic reasoning profile in under
500 verified problems.

Our main contributions are:
\begin{itemize}
  \item \textbf{Nine-Dimension Algebraic Complexity Framework.} The first
  framework to define and jointly operationalise nine orthogonal complexity
  dimensions for algebraic reasoning, each traced to prior literature on LLM
  failure modes.
  \item \textbf{Automated Generation and Verification Pipeline.} We present a parametric system that produces CAS-verified problems at any desired complexity level. The system operates across all nine dimensions without requiring any human annotation. This makes it a dynamic benchmark. It remains relevant even as model capabilities continue to improve.
  \item \textbf{Comprehensive Cross-Model Evaluation.} We compute per-dimension failure curves across seven models ranging from 8B to 235B parameters. These curves identify precise failure thresholds for each dimension. They also allow us to isolate universal architectural limits from model-specific performance gaps. From this analysis, we derive a five-dimension diagnostic shortlist that is well-suited for algebraic complexity profiling.
\end{itemize}

%%, ------------------------------------------------------------
\section{Related Work}

\subsection{Algebraic Reasoning in LLMs}

Standard mathematical reasoning benchmarks, 
GSM8K~\citep{cobbe2021gsm8k}, MATH~\citep{hendrycks2021measuring},
HardMath~\citep{fan2024hardmath}, report a single accuracy score over
heterogeneous problem sets. \citet{song2026} synthesise over 200 papers into a
taxonomy of LLM reasoning failures. The shared limitation is that expression
depth, operator type, operand size, and branching structure all co-vary across
problems. And it makes it impossible to attribute failure to any one cause. Individual
failure modes have been studied in isolation. \citet{lample2020deep} showed neural
models collapse as operators grow toward nested transcendental operators.
\citet{saxton2019analysing} found division and integration at the bottom of every
accuracy chart. \citet{dziri2024faith} gave the mechanistic account, GPT-4
drops from 59\% on three-digit multiplication to 4\% on four-digit due to
$O(n^2)$ cross-digit interactions. And \citet{sander2024} traced this to
carry-cascade complexity. \citet{zhao2024} and \citet{hosseini2024} showed that the
compositionality breaks when sub-problems are chained together. Counting
failures originate in tokenisation and positional encoding
limits~\citep{zhang2024tokenization,chang2024} a finding \citet{malek2025}
confirmed in reasoning-specialised models. \citet{markeeva2024clrs} is the
closest methodological precedent, they had varied trace length and problem size as
separate axes on CLRS-Text algorithmic tasks. Our work extends that principle to nine dimensions jointly within an algebraic domain. We evaluate across seven models of varying scale. To our knowledge, this is the first controlled evaluation that spans all documented failure modes simultaneously.

\subsection{Automated Generation of Mathematical Problems}

The earliest principled work on algebraic problem generation is
\citet{singh2012algebra}. It generates problem variants through syntactic
generalisation and verifies correctness using polynomial identity testing producing provably valid problems across polynomials, trigonometry, calculus, and
determinants. \citet{xu2021procedural} describe a template-based procedural system
that creates abstract problems at varying difficulty levels and realises them in
natural language. They report 56\% time savings over manual creation. More recent work has shifted toward LLM-assisted problem generation at training scale. MathScale~\citep{tang2024mathscale} constructs a concept graph from seed questions. It then uses frontier LLMs to produce two million question-answer pairs. This approach yields a 43\% improvement in macro accuracy. SAND-Math~\citep{manem2025sandmath} introduces a Difficulty Hiking
pipeline that generates problems and systematically elevates their complexity,
boosting AIME25 performance by 17.85 points over the next-best synthetic dataset.
\citet{chen2025arrows} generate executable programs encoding math problems. Then
translate them to natural language, and validate answers bilaterally against
program outputs across 12.3 million triples. \citet{ariyarathne2025wordproblems}
find that LLM-generated word problems are generally high quality but that models
still struggle to reliably match specified grade levels. All of this prior work
generates problems primarily as training data, with difficulty either hand-tuned
or LLM-estimated. Our system is designed for diagnostic evaluation rather than training data generation. Each generator varies exactly one complexity dimension while holding the remaining eight dimensions fixed. Every problem is CAS-verified before it is presented to any model. This process produces isolation-controlled accuracy curves instead of training corpora.

%%, ------------------------------------------------------------
\section{Benchmark Design and Methodology}

Standard benchmarks collapse every source of difficulty into one accuracy score.
Existing benchmarks do not isolate the cause of model failures. Our benchmark does. The design rests on two core principles. The first is to identify every structurally distinct dimension along which an algebraic expression can become harder. The second is to build a generation pipeline that varies each dimension in strict isolation while holding the remaining eight fixed. The result is fully
parametric, complexity levels are arguments to a generator, not fixed features
of a static corpus, so the framework stays relevant as models improve.

\subsection{Nine Dimensions of Algebraic Complexity}

Each dimension corresponds to one documented failure mode in the LLM reasoning
literature, with a mechanistic explanation and prior empirical grounding.All problems are represented in Polish prefix notation. This format makes tree depth, branching, and token count directly readable from the token sequence without any additional parsing.

\paragraph{D1, Syntactic Length.}
\textbf{Definition:} Total token count in prefix notation. \citet{markeeva2024clrs}
showed transformers do not reliably extrapolate to lengths beyond their training
distribution. Also the positional encodings accumulate error at each additional token
even when individual steps are trivial. We built Nine levels from 5 to 751 tokens,
as right-growing addition chains where token count follows $2n{-}1$ for $n$ terms.

\textbf{Examples:} \texttt{3+2+6} (5 tokens) $\to$ \texttt{2+3+4+9+...}
(751 tokens).

\paragraph{D2, Tree Depth.}
\textbf{Definition:} Longest root-to-leaf path in the expression tree. At depth
$d$, a model must hold $2^d$ partial results simultaneously before the root
operation resolves which is an exponential scaling of concurrent memory demand.
\citet{saxton2019analysing} quantified ${\sim}15\%$ accuracy loss per additional
nesting level. Also \citet{dziri2024faith} gave the mechanistic account. Problems use
right-spine trees so only nesting depth varies.

\textbf{Examples:} \texttt{3+2} (depth 1) $\to$
\texttt{3+7*(6+5*(2+5*(3+3*4)))} (depth 6).

\paragraph{D3, Operator Score ($\sigma$).}
\textbf{Definition:} Ordinal hardness rank per operator, ordered
\texttt{neg} $<$ \texttt{abs} $<$ \texttt{add} $<$ \texttt{sub} $<$
\texttt{mul} $<$ \texttt{div} $<$ \texttt{sqrt} $<$ \texttt{exp} $<$
\texttt{ln} $<$ \texttt{sin/cos} $<$ \texttt{tan} $<$ \texttt{pow}. The ranking
is derived from training data frequency~\citep{lample2020deep,biggio2021},
cross-token interaction complexity ($O(n^2)$ for \texttt{mul}/\texttt{pow}.
\citealt{dziri2024faith}), and polynomial approximation degree for
transcendentals (\texttt{tan} $\approx$ degree 27 vs \texttt{exp} $\approx$
degree 10; \citealt{fog2025instruction}).

We even validated it on Qwen-2.5-7B-Instruct with all structural dimensions held at minimum
so operator identity was the sole varying factor. The ordering achieved Spearman
$\rho = 0.863$, stable across two independent runs. One anomaly: \texttt{ln}
scored 100\% despite its mid-hard rank ($\sigma{=}15$). It happened most plausibly because
the well-formed \texttt{ln(x)} prompts trigger lookup-style retrieval of memorised
identities (\texttt{ln(1)=0}, \texttt{ln(e)=1}) rather than genuine computation.
\citet{weber2002} documents the same pattern in students. The rank is retained at
$\sigma{=}15$; human error rates on logarithm problems
(40--60\%; \citealt{weber2002}) propagate difficulty into the training signal
regardless.

\textbf{Examples:} \texttt{-3} (neg, $\sigma{=}2$) $\to$ \texttt{21/3}
($\sigma{=}9$) $\to$ \texttt{2\^{}7} (pow, $\sigma{=}22$).

\paragraph{D4, Working Memory.}
\textbf{Definition:} Count of parallel independent sub-results that must coexist
before any can be combined. \citet{gong2023working} showed formally that
self-attention limits working memory capacity, there is no register file. Also,
holding many independent values concurrently is outside what attention was
designed to do. Each problem is a sum of $K$ independent single-digit products.
The products are individually trivial; the only challenge is tracking $K$ results
while waiting to sum them. Levels: 2 to 200 parallel branches.

\textbf{Examples:} \texttt{3*2+6*5} ($K{=}2$) $\to$
\texttt{6*7+5*3+8*4+...} ($K{=}30$ parallel products).

\paragraph{D5, Compositional Branching.}
\textbf{Definition:} Operations inside each branch of a two-branch tree before
the branch resolves to a scalar. D4 counts how many branches coexist (width of the tree); D5 counts
how deep each branch goes (depth of the tree). \citet{zhao2024} established that compositionality breaks
specifically when sub-problems are chained; \citet{hosseini2024} replicated this
across model families. Problems always have two branches; ops per branch scales
0 to 30.

\textbf{Examples:} \texttt{3+2+6} (0 ops/branch) $\to$
\texttt{2*(9+9*(2+2+8+...))} (30 ops/branch).

\paragraph{D6, Solution Ambiguity.}
\textbf{Definition:} Number of structurally distinct valid solution strategies
for a problem. When multiple valid paths exist, a model must implicitly select
one. This raises search cost before any arithmetic begins. \citet{lample2020deep}
found beam search became essential whenever equivalent-length solutions existed.
Eight problem types ordered from linear equations (one path) to algebraic
identity simplification (four or more paths).

\textbf{Examples:} \texttt{3x+6=-9} (1 path) $\to$
\texttt{(8\^{}2-2\^{}2)/(8+2)} (4+ paths).

\paragraph{D7, Counting Load ($K$).}
\textbf{Definition:} Number of identical repeated operands requiring explicit
enumeration. \citet{song2026} describe counting as a fundamental architectural
challenge; \citet{malek2025} confirmed it persists in reasoning-specialised
models. The failure is rooted in tokenisation, identical repeated tokens lack
positionally distinct representations~\citep{zhang2024tokenization,chang2024}.
Levels: $K = 5$ to 300 number of digits. 

\textbf{Examples:} \texttt{7+7+7+7+7} ($K{=}5$) $\to$ \texttt{5+5+5+...}
($K{=}300$).

\paragraph{D8, Sequential Chain Length.}
\textbf{Definition:} Length of the longest strictly dependent step path, where
each result feeds directly into the next and only one live intermediate value
exists at any step. Unlike D4 where results accumulate in parallel, here a single
result is passed forward sequentially. The failure mode is error compounding
rather than memory overflow, \citet{merrill2023sequential} showed each
additional hop multiplicatively increases computational demand. Problems are
generated as left-spine chains. At each step, an operator from
$\{+,\,-,\,\times\}$ is applied to the running result along with a small operand. A
value-bounds guard ($-50\text{k}$ to $500\text{k}$, non-zero) prevents numeric
blowup. Difficulty levels range from 1 to 12 steps.

\textbf{Examples:} \texttt{3+4} (1 step) $\to$
\texttt{(((2+3+5-5+3)*3+3-3+6)*2...)} (12 dependent steps).

\paragraph{D9, Numeric Magnitude.}
\textbf{Definition:} Maximum digit count across all operands. With addition,
$O(n)$ carry propagation scales gracefully. With multiplication, $O(n^2)$
partial-product accumulation creates cross-digit interactions that outpace
attention. \citet{yuan2023} showed sharp accuracy drops as operand size grows.
All D9 problems use multiplication to keep the quadratic interaction active.
Digit counts: 1, 2, 4, 6, 8, 15.

\textbf{Examples:} \texttt{3*2} (1-digit) $\to$
\texttt{492950566229566 * 177454928531585} (15-digit).

\subsection{Automated Generation and Verification}

Most benchmarks found in the literature are static, the problem set is fixed at collection time. And
once frontier models saturate it, the benchmark loses diagnostic value. Our
generation system avoids this. Every dimension is fully parametric and random seed are arguments to a generator script. Extending a
suite to a higher level requires changing one integer. New suites for future
dimensions require only a new generator module that conforms to the shared
interface.

% \subsubsection*{Expression representation}

% Algebraic expressions are naturally tree-structured, and we use Polish prefix
% notation to represent them. In prefix form, tree depth, branching factor, and
% token count are directly readable from the token sequence without any parsing ---
% a property that makes generation constraints expressible as token-level rules.
% Each generator constructs prefix strings directly: D1 builds right-growing
% addition chains, D4 right-nests independent products under a sum, and D8 builds
% left-spine chains by wrapping the running result at each step.

\subsubsection*{Dimension isolation}

Each generator varies exactly one parameter while fixing the other eight at
minimum values. D1 uses flat right-spine addition chains with single-digit
operands, varying term count from 3 to 376 (5 to 751 tokens). D2 uses
right-spine trees with \texttt{add}/\texttt{mul} operators, varying nesting depth
from 1 to 8. D4 sums $K$ independent single-digit products, varying $K$ from 2
to 200. D7 sums one fixed integer repeated $K$ times ($K{=}5$ to 300), paired
with a control that encodes the same result as base-$K$ exponentiation to isolate
counting failure from arithmetic difficulty. D8 builds left-spine chains with
operators drawn from $\{+,\,-,\,\times\}$ at each step, varying chain length from
1 to 12, with a value-bounds guard to prevent numeric blowup. For D3, all
structural dimensions are at minimum and only the operator changes across twelve
levels following the $\sigma$-rank ordering.

\subsubsection*{CAS verification}

Every problem is passed through SymPy before evaluation. Arithmetic, equations, derivatives, integrals, and modular operations each have dedicated verification handlers. Each handler operates under a 12-second timeout. Problems that time out or raise solver exceptions are
discarded and replaced; only those with clean, finite answers enter the final set.

\subsubsection*{Model evaluation protocol}

Before reaching any model, prefix expressions are converted to ASCII infix format. This is the standard way humans write algebra and ensures consistent tokenisation across all model vocabularies. For example, the prefix expression \texttt{mul add 3 2 sub 6 5} becomes \texttt{(3 + 2) * (6 - 5)}.
Unicode math symbols are replaced with plain ASCII (\texttt{*}, \texttt{\^{}}). All seven models are
queried via API with a fixed system prompt requiring step-by-step solving and a
final answer on a dedicated \texttt{ANSWER: <value>} line, at temperature zero for
deterministic outputs. Each complexity level contains 50 problems generated with
random seed~42. The correctness is judged at $\pm$0.5\% relative or $\pm$0.05
absolute tolerance. The full pipeline, generation, CAS verification, format
conversion, and LLM evaluation, is end-to-end automated with no human
annotation at any stage.

%%, ------------------------------------------------------------
\section{Experiments and Results}

\begin{figure}[t]
  \centering
  \includegraphics[width=\linewidth]{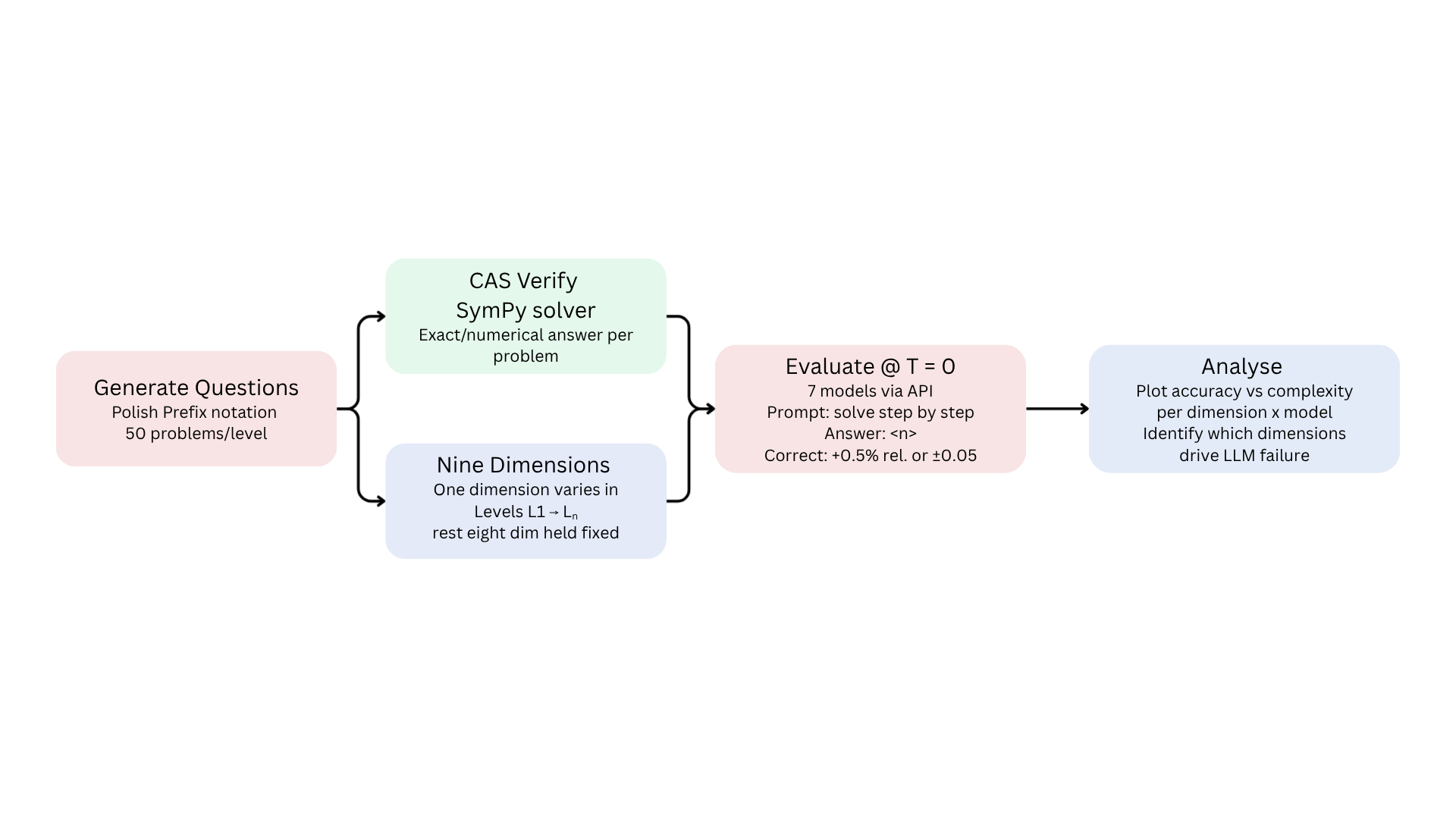}
  \caption{Experimental pipeline. Problems are generated in Polish prefix notation, verified
  by a SymPy CAS, converted to ASCII infix, and evaluated across seven models at temperature
  zero.}
  \label{fig:pipeline}
\end{figure}

We evaluated seven instruction-tuned models, GPT-4o Mini, Claude 3.5 Haiku, Qwen3 235B,
DeepSeek V3, Gemma 3 12B, Ministral 8B, and Llama 3 8B, across all nine batteries. Each
model was queried at temperature zero via API. No existing benchmark was reused.
Figure~\ref{fig:pipeline} illustrates the end-to-end experimental pipeline: problems are
generated in Polish prefix notation, verified by a SymPy CAS, converted to ASCII infix, and
evaluated across all seven models.

\begin{figure}[htbp]
  \centering
  \includegraphics[width=\linewidth]{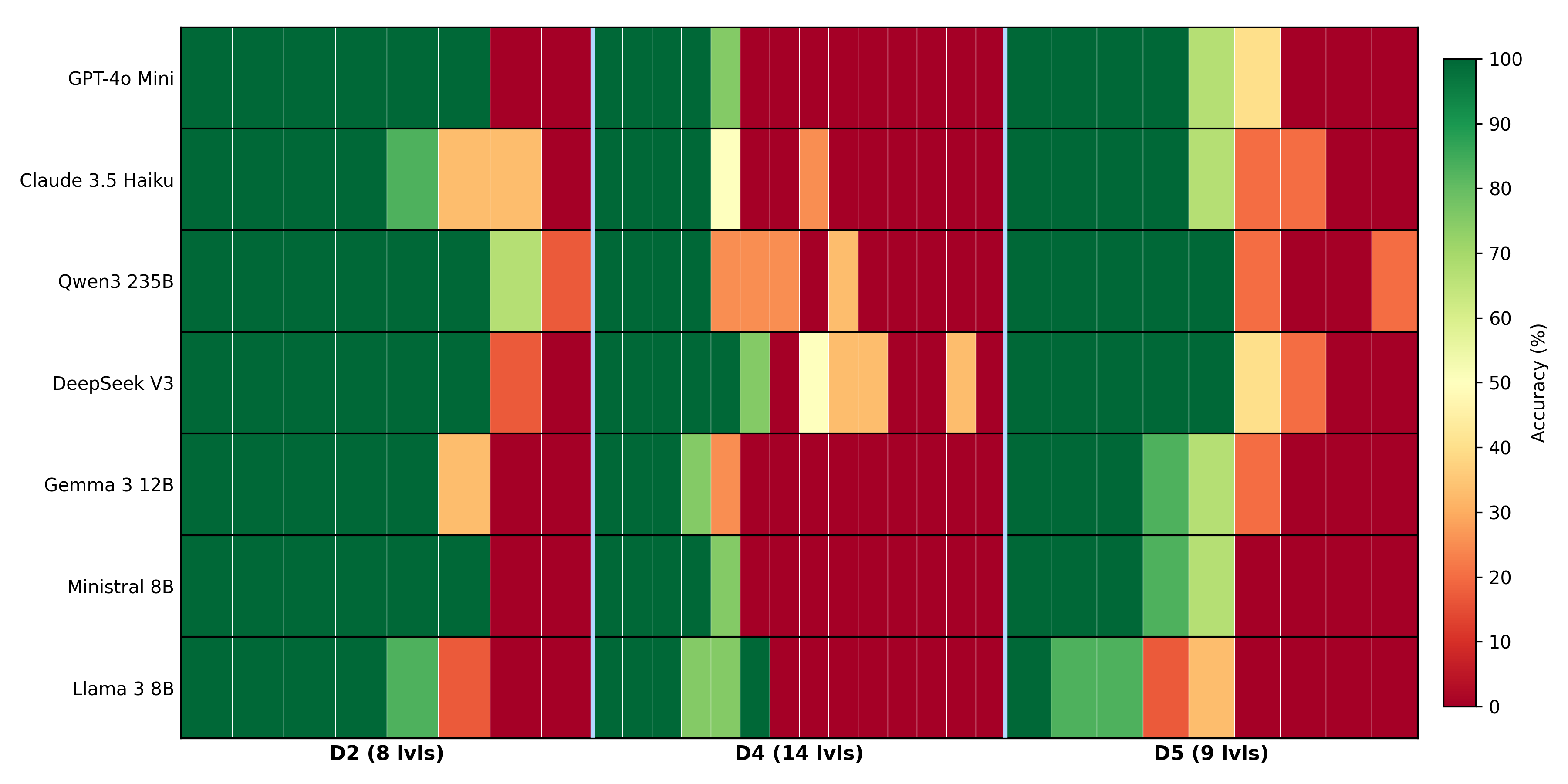}
  \caption{Accuracy heatmaps for D4 (14 levels, 2--200 parallel branches), D2 (8 levels,
  depth 1--8), and D5 (9 levels, 0--30 ops/branch). Rows = 7 models; columns = increasing
  complexity; colour: dark green $\approx$ 100\%, red = 0\%.}
  \label{fig:D2D4D5nested_heatmap}
\end{figure}

\subsection{Core Structural Predictors: D4, D2, D5 (Figure~\ref{fig:D2D4D5nested_heatmap})}

\textbf{D4, Working memory.}
All seven models score 100\% through 12 parallel branches and then they collapse abruptly between 20
and 30. Qwen3 235B and Llama 3 8B fail at the same threshold despite a 30$\times$ parameter
gap. We observe here that for working memory, scale offers no rescue whatsoever. Transformer attention has no dedicated register
mechanism for holding $K$ co-existing intermediate values simultaneously, this is an
architectural constraint, not a capacity limit that training data or parameter count can
address~\citep{markeeva2024clrs,gong2023working}.

\textbf{D2, Tree depth.}
Qwen3 235B is the last to fail, retaining 17\% at depth 8. Its larger parameter count helps track nested partial results longer. However, size is not the deciding factor. Claude 3.5 Haiku outperforms Deepseek V3 despite having far fewer parameters. Since the two best-performing models follow entirely different architectural approaches, the advantage is unlikely to be architectural. Training data quality, reasoning supervision, or instruction tuning are more plausible explanations. 

\textbf{D5, Compositional branching.}
All models hold at 100\% through 3 ops/branch. Llama 3 8B is the first to break, collapsing
at exactly 3 ops/branch (17\%), the earliest model-specific failure point in the entire
study. Claude 3.5 Haiku is the most resilient, holding 20\% at 12 ops/branch where all
others are at 0\%. The key finding is the orthogonality with D4: Llama 3 8B handles parallel
branch count normally in D4 but fails almost immediately in D5. Failure in D5 is driven by local sub-expression depth within each branch. The number of coexisting branches is not the determining factor~\citep{zhao2024,song2026}.

\begin{figure}[htbp]
  \centering
  \includegraphics[width=\linewidth]{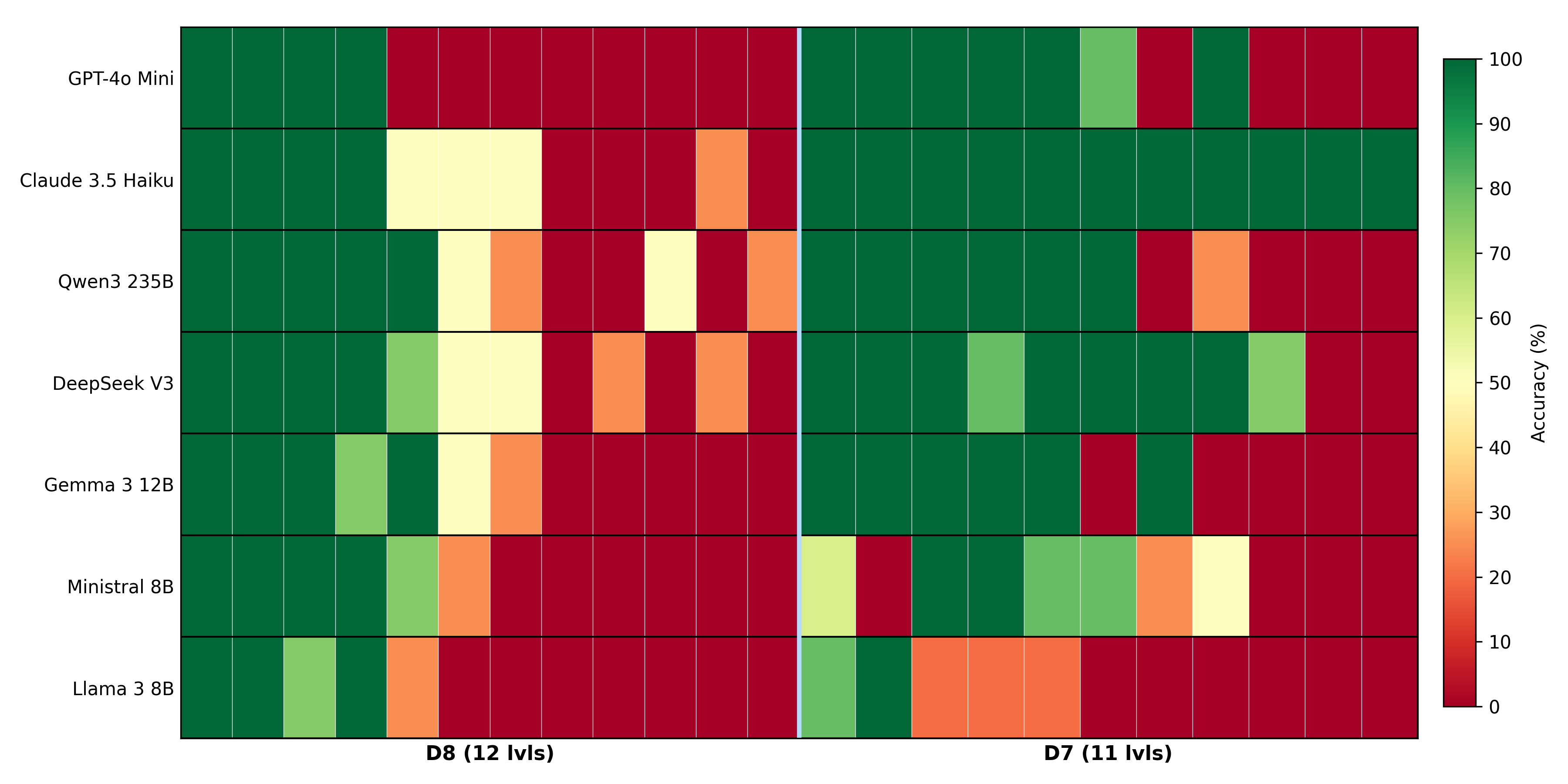}
  \caption{Accuracy heatmaps for D8 (12 levels, chain steps 1 to 12) and D7 (11 levels,
  $K$ = 5 to 300 identical terms). Same colour encoding as
  Figure~\ref{fig:D2D4D5nested_heatmap}.}
  \label{fig:D8D7nested_heatmap}
\end{figure}

\subsection{Sequential Chaining and Counting: D8 and D7 (Figure~\ref{fig:D8D7nested_heatmap})}

\textbf{D8, Sequential chain length.}
GPT-4o Mini shows a step-function collapse: 100\% at steps 1to4, then a hard drop to 0\%
from step 5 with no intermediate degradation anywhere, the sharpest single-step threshold
in the study. Most models fail by step 7; none retain any accuracy at step 12. This substantially extends earlier findings. Prior studies were limited to two models and capped at 9 steps. Those studies classified D8 as only moderately destructive. The last three levels reveal total failure, making D8 as
catastrophic as D2 at sufficient depth. Each additional dependent step multiplicatively
compounds error~\citep{merrill2023sequential}, and that product eventually reaches one.

\textbf{D7, Counting load.}
This dimension shows the widest model divergence of any dimension. Claude 3.5 Haiku scores
100\% at every $K$ from 5 to 300. It does this by recognising repeated addition as multiplication and computing
$K \times \text{value}$ directly, bypassing the tokenisation ceiling that causes other models
to fail. Llama 3 8B fails from $K{=}25$; Gemma 3 12B alternates between passing and failing,
indicating inconsistent strategy use. A paired control suite confirms that the failure is in cardinality tracking rather than arithmetic ability. The control problems use identical arithmetic through base-K exponentiation but do not require any counting. A model that passes the control but fails the primary problem at the same K value is provably failing at counting~\citep{zhang2024tokenization,chang2024}.

\begin{figure}[htbp]
  \centering
  \includegraphics[width=\linewidth]{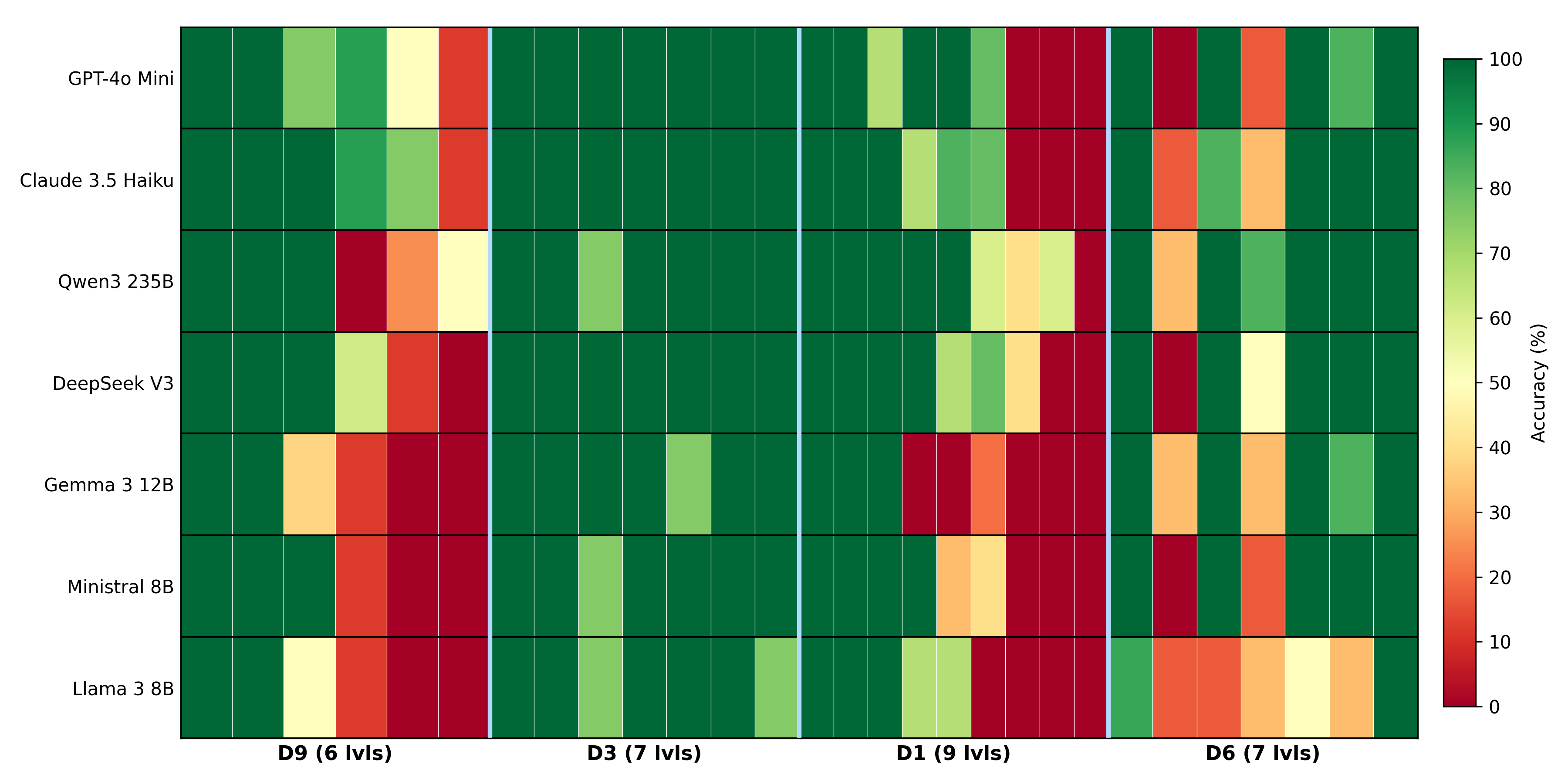}
  \caption{Accuracy heatmaps for D9 (6 levels, 1--15 digit operands), D3 (12 levels,
  $\sigma = 2$--22), D1 (9 levels, 5--751 tokens), and D6 (8 problem types). Same colour
  encoding as Figure~\ref{fig:D2D4D5nested_heatmap}.}
  \label{fig:D9D3D1D6_nested_heatmap}
\end{figure}

\subsection{Secondary Dimensions: D9, D3, D1, D6 (Figure~\ref{fig:D9D3D1D6_nested_heatmap})}

\textbf{D9} shows a family-level split, frontier models hold above 50\% through 8-digit
multiplication, whereas the smaller models collapse at 4 digits. And all fail at 15 digits, but
difficulty is conditional on operator. D9 amplifies D3: large operands are manageable under
addition ($O(n)$ carry) but intractable under multiplication ($O(n^2)$ partial products). It
is not an independent predictor~\citep{dziri2024faith}. \textbf{D3} is near-100\% across all
12 operator levels; minor dips appear only at \texttt{exp} and \texttt{pow-hard} for weaker
models. The catastrophic failure zone begins above the $\sigma{=}22$ ceiling, This is calculus-level
complexity and is not covered here~\citep{lample2020deep,saxton2019analysing}. \textbf{D1} collapses
universally above 201 tokens, but this is a downstream effect of D2 and D8, long
expressions are necessarily deeply nested or long chains. Non-monotonic mid-range accuracy
confirms D1 is a proxy variable, not a causal factor~\citep{markeeva2024clrs}. \textbf{D6}
shows no monotonic pattern; quadratic factoring is hard everywhere, algebraic identities easy
everywhere.

\subsection{Correlations and Interactions Across Dimensions}

The nine dimensions are actually not fully independent. Several pairs exhibit interpretable
correlations. Like, D1 and D8 are strongly correlated at extreme lengths because a 751-token
expression almost always encodes a long sequential chain. D1 thus acts as a noisy
proxy for D8, to a lesser extent, D2. So it is not an independent predictor. D9 and D3
interact by design. Large numeric magnitude only becomes a genuine source of failure when the operator
is multiplication. This makes D9 an amplifier of the D3 signal rather than a standalone
dimension. D4 and D5 both involve branching yet capture orthogonal constraints. Llama 3 8B performs near-normally on D4 but collapses catastrophically on D5. This sharp contrast is the clearest empirical confirmation that the two dimensions are genuinely independent. D2 and D4 share
working memory as an underlying resource but stress it differently: D2 demand grows
exponentially with nesting depth ($2^d$ concurrent results), while D4 demand grows linearly
with branch count. A model can therefore fail D2 at depth 7 while still handling D4 at 12
branches, the two curves are not interchangeable.D7 is the most orthogonal of all nine dimensions. A model can pass D2, D4, D5, and D8 at moderate difficulty levels while simultaneously failing D7 at K=30. This is because counting failure is rooted in tokenisation rather than tree structure or memory capacity~\citep{zhang2024tokenization,chang2024}.

\subsection{Five Dimensions Are Diagnostically Sufficient}

The nine-suite evaluation shows that five dimensions, D2, D4, D5, D7, D8, jointly
cover the full space of documented LLM algebraic failure modes, each capturing a
mechanistically distinct bottleneck. D4 covers horizontal memory overflow from co-existing
intermediates~\citep{markeeva2024clrs,gong2023working}. D2 covers exponential vertical nesting
demand~\citep{saxton2019analysing,dziri2024faith}. D5 covers local sub-expression depth,
confirmed as orthogonal to D4 by Llama 3 8B's divergent profiles~\citep{zhao2024,song2026}.
D8 covers sequential error compounding, shown here to be as severe as D2 at sufficient chain
length~\citep{merrill2023sequential}. D7 covers tokenisation-level cardinality limits, entirely
orthogonal to the four structural dimensions~\citep{zhang2024tokenization,shin2024kaneko}. The
remaining four are derivable or subsumed: D1 is a proxy for D2$+$D8; D6 is subsumed by D3;
D9 belongs inside a D3$\times$D9 interaction term. D3 is the most important exclusion, its
catastrophic floor lies above the current suite ceiling and should be probed with
calculus-level operators in any complete diagnostic framework. A suite spanning D2, D4, D5,
D7, D8, and extended D3 characterises a model's full algebraic reasoning profile. Each dimension is a one-parameter change to the generator, so the framework can be extended to higher difficulty levels as models improve without any redesign. And 250 problems (50 per dimension) are sufficient for a complete diagnostic profile, making it practical for regular use in model development cycles.

%%, ------------------------------------------------------------
\section{Conclusion}

We introduced a parametric benchmark of nine independently tested algebraic complexity
dimensions and evaluated seven models across all of them. Working memory is a scale-invariant architectural limit. Every model collapses at 20 to 30 parallel branches regardless of parameter count. Sequential chaining is
more destructive than prior studies reported; extending the suite from 9 to 12 steps reveals
total failure where earlier work saw only partial degradation. Counting failure is orthogonal
to all structural failure modes: a model's profile on D2, D4, D5, and D8 predicts almost
nothing about its D7 performance, and the gap between Claude 3.5 Haiku (100\% at $K{=}300$)
and Llama 3 8B (fails at $K{=}25$) is invisible in any aggregate accuracy score. The
five-dimension shortlist gives a complete diagnostic picture, and
because the benchmark is fully generative, raising the complexity ceiling on any dimension is
a one-parameter change, the framework stays relevant as models improve. Future research
will focus on whether fine-tuning on data generated through this framework improves model
accuracy in algebraic problem-solving. Additionally, we plan to examine how LLM performance
across these complexity dimensions compares with human cognitive behaviour and failure
patterns.
\section*{Acknowledgements}

We thank the reviewers for their thoughtful and constructive feedback. We also thank our colleagues for helpful discussions during the development of this study.

The authors acknowledge the use of AI-assisted writing tools, including Claude and Gemini, during the preparation of this manuscript. These tools were used solely to improve the clarity, presentation, and grammatical quality of the writing. All experimental results, analyses, conclusions, and proposed techniques are entirely the authors' own work and remain a concrete representation of their intellectual contributions. The authors take full responsibility for the contents of this paper.

\newpage

\bibliography{colm2026_conference}

\begin{thebibliography}{27}
\providecommand{\natexlab}[1]{#1}
\providecommand{\url}[1]{\texttt{#1}}
\expandafter\ifx\csname urlstyle\endcsname\relax
  \providecommand{\doi}[1]{doi: #1}\else
  \providecommand{\doi}{doi: \begingroup \urlstyle{rm}\Url}\fi

\bibitem[Ariyarathne et~al.(2025)]{ariyarathne2025wordproblems}
Gihan Ariyarathne et~al.
\newblock Evaluating {LLM}-generated mathematical word problems: Quality and grade-level alignment.
\newblock \emph{arXiv preprint}, 2025.

\bibitem[Biggio et~al.(2021)Biggio, Bendinelli, Neitz, Lucchi, and Parascandolo]{biggio2021}
Luca Biggio, Tommaso Bendinelli, Alexander Neitz, Aurelien Lucchi, and Giambattista Parascandolo.
\newblock Neural symbolic regression that scales.
\newblock In \emph{Proceedings of the 38th International Conference on Machine Learning}, volume 139 of \emph{PMLR}, pp.\  936--945, 2021.

\bibitem[Chang \& Bisk(2024)Chang and Bisk]{chang2024}
Yingshan Chang and Yonatan Bisk.
\newblock Language models need inductive biases to count inductively.
\newblock \emph{arXiv preprint arXiv:2405.20131}, 2024.

\bibitem[Chen et~al.(2025)Chen, Tian, Hu, Chen, Liu, Zhang, and Zhou]{chen2025arrows}
Sheng Chen, Chao Tian, Bo~Hu, Kang Chen, Zheng Liu, Zheng Zhang, and Ji~Zhou.
\newblock Arrows of math reasoning data synthesis for large language models: Diversity, complexity and correctness.
\newblock In \emph{Proceedings of the 34th {ACM} International Conference on Information and Knowledge Management}, pp.\  4665--4669, 2025.

\bibitem[Cobbe et~al.(2021)Cobbe, Kosaraju, Bavarian, Chen, Jun, Kaiser, Plappert, Tworek, Hilton, Nakano, et~al.]{cobbe2021gsm8k}
Karl Cobbe, Vineet Kosaraju, Mohammad Bavarian, Mark Chen, Heewoo Jun, Lukasz Kaiser, Matthias Plappert, Jerry Tworek, Jacob Hilton, Reiichiro Nakano, et~al.
\newblock Training verifiers to solve math word problems.
\newblock \emph{arXiv preprint arXiv:2110.14168}, 2021.

\bibitem[Dziri et~al.(2024)Dziri, Lu, Sclar, Li, Jiang, Lin, Welleck, West, Bhagavatula, Le~Bras, et~al.]{dziri2024faith}
Nouha Dziri, Ximing Lu, Melanie Sclar, Xiang~Lorraine Li, Liwei Jiang, Bill~Yuchen Lin, Sean Welleck, Peter West, Chandra Bhagavatula, Ronan Le~Bras, et~al.
\newblock Faith and fate: Limits of transformers on compositionality.
\newblock In \emph{Advances in Neural Information Processing Systems}, volume~36, 2024.

\bibitem[Fan et~al.(2024)]{fan2024hardmath}
Jingxuan Fan et~al.
\newblock {HardMath}: A benchmark dataset for challenging problems in applied mathematics.
\newblock \emph{arXiv preprint}, 2024.

\bibitem[Fog(2025)]{fog2025instruction}
Agner Fog.
\newblock Instruction tables: Lists of instruction latencies, throughputs and micro-operation breakdowns for {Intel}, {AMD} and {VIA} {CPUs}.
\newblock \url{https://www.agner.org/optimize/}, 2025.

\bibitem[Gong \& Zhang(2024)Gong and Zhang]{gong2023working}
Dongyu Gong and Hantao Zhang.
\newblock Self-attention limits working memory capacity of transformer-based models.
\newblock \emph{arXiv preprint arXiv:2409.10715}, 2024.

\bibitem[Hendrycks et~al.(2021)Hendrycks, Burns, Kadavath, Arora, Basart, Tang, Song, and Steinhardt]{hendrycks2021measuring}
Dan Hendrycks, Collin Burns, Saurav Kadavath, Akul Arora, Steven Basart, Eric Tang, Dawn Song, and Jacob Steinhardt.
\newblock Measuring mathematical problem solving with the {MATH} dataset.
\newblock \emph{arXiv preprint arXiv:2103.03874}, 2021.

\bibitem[Hosseini et~al.(2024)]{hosseini2024}
Arian Hosseini et~al.
\newblock Not all {LLM} reasoners are created equal.
\newblock \emph{arXiv preprint}, 2024.

\bibitem[Lample \& Charton(2020)Lample and Charton]{lample2020deep}
Guillaume Lample and Fran\c{c}ois Charton.
\newblock Deep learning for symbolic mathematics.
\newblock In \emph{International Conference on Learning Representations (ICLR)}, 2020.

\bibitem[Malek et~al.(2025)Malek, Ge, Lazic, Jin, Gy{\"o}rgy, and Szepesv{\'a}ri]{malek2025}
Alan Malek, Jiawei Ge, Nevena Lazic, Chi Jin, Andr{\'a}s Gy{\"o}rgy, and Csaba Szepesv{\'a}ri.
\newblock Frontier {LLMs} still struggle with simple reasoning tasks.
\newblock \emph{arXiv preprint arXiv:2507.07313}, 2025.

\bibitem[Manem et~al.(2025)]{manem2025sandmath}
Venkat~Srinivasan Manem et~al.
\newblock {SAND-Math}: Difficulty hiking for mathematical reasoning.
\newblock \emph{arXiv preprint}, 2025.

\bibitem[Markeeva et~al.(2024)]{markeeva2024clrs}
Larisa Markeeva et~al.
\newblock The {CLRS-Text} algorithmic reasoning language benchmark.
\newblock In \emph{Proceedings of the 41st International Conference on Machine Learning (ICML)}, 2024.

\bibitem[Merrill \& Sabharwal(2023)Merrill and Sabharwal]{merrill2023sequential}
William Merrill and Ashish Sabharwal.
\newblock The expressive power of transformers with chain of thought.
\newblock \emph{arXiv preprint arXiv:2310.07923}, 2023.

\bibitem[Sander et~al.(2024)Sander, Giryes, Suzuki, Blondel, and Peyr{\'e}]{sander2024}
Micha{\"e}l~E. Sander, Raja Giryes, Taiji Suzuki, Mathieu Blondel, and Gabriel Peyr{\'e}.
\newblock How do transformers perform in-context autoregressive learning?
\newblock In \emph{Proceedings of the 41st International Conference on Machine Learning (ICML)}, volume 235 of \emph{PMLR}, 2024.

\bibitem[Saxton et~al.(2019)Saxton, Grefenstette, Hill, and Kohli]{saxton2019analysing}
David Saxton, Edward Grefenstette, Felix Hill, and Pushmeet Kohli.
\newblock Analysing mathematical reasoning abilities of neural models.
\newblock In \emph{International Conference on Learning Representations (ICLR)}, 2019.

\bibitem[Shin \& Kaneko(2024)Shin and Kaneko]{shin2024kaneko}
Andrew Shin and Katsuhito Kaneko.
\newblock Large language models lack understanding of character composition of words.
\newblock \emph{arXiv preprint arXiv:2405.11357}, 2024.

\bibitem[Singh et~al.(2012)Singh, Gulwani, and Rajamani]{singh2012algebra}
Rishabh Singh, Sumit Gulwani, and Sriram~K. Rajamani.
\newblock Automatically generating algebra problems.
\newblock In \emph{Proceedings of the 26th {AAAI} Conference on Artificial Intelligence}, pp.\  1620--1628, 2012.
\newblock \doi{10.1609/aaai.v26i1.8341}.

\bibitem[Song et~al.(2026)Song, Han, and Goodman]{song2026}
Peiyang Song, Pengrui Han, and Noah Goodman.
\newblock Large language model reasoning failures.
\newblock \emph{Transactions on Machine Learning Research}, January 2026.

\bibitem[Tang et~al.(2024)Tang, Zhang, Wang, and Wei]{tang2024mathscale}
Zhengyang Tang, Xingxing Zhang, Benyou Wang, and Furu Wei.
\newblock {MathScale}: Scaling instruction tuning for mathematical reasoning.
\newblock \emph{arXiv preprint arXiv:2403.02884}, 2024.

\bibitem[Weber(2002)]{weber2002}
Keith Weber.
\newblock Students' understanding of exponential and logarithmic functions.
\newblock \emph{International Conference on the Teaching of Mathematics}, 2002.

\bibitem[Xu et~al.(2021)Xu, Smeets, and Bidarra]{xu2021procedural}
Yi~Xu, Roger Smeets, and Rafael Bidarra.
\newblock Procedural generation of problems for elementary math education.
\newblock \emph{International Journal of Serious Games}, 8\penalty0 (2), 2021.
\newblock \doi{10.17083/ijsg.v8i2.396}.

\bibitem[Yuan et~al.(2023)]{yuan2023}
Zheng Yuan et~al.
\newblock How well do large language models perform in arithmetic tasks?
\newblock \emph{arXiv preprint}, 2023.

\bibitem[Zhang et~al.(2024)Zhang, Cao, and You]{zhang2024tokenization}
Xingwei Zhang, Jiahui Cao, and Chengyuan You.
\newblock Counting ability of large language models and impact of tokenization.
\newblock \emph{arXiv preprint arXiv:2410.19730}, 2024.

\bibitem[Zhao et~al.(2024)Zhao, Tong, Mou, Zhang, Zhang, and Huang]{zhao2024}
Jun Zhao, Jingqi Tong, Yurong Mou, Ming Zhang, Qi~Zhang, and Xuanjing Huang.
\newblock Exploring the compositional deficiency of large language models in mathematical reasoning through trap problems.
\newblock In \emph{Proceedings of the 2024 Conference on Empirical Methods in Natural Language Processing (EMNLP)}, pp.\  16361--16376, 2024.

\end{thebibliography}
\bibliographystyle{colm2026_conference}

\appendix

\section{Per-dimension accuracy results}
\label{app:results}

\renewcommand{\thefigure}{A\arabic{figure}}
\setcounter{figure}{0}

All nine suite results are presented as line graphs. Each line corresponds to
one of the seven models, traced across increasing complexity levels from left to
right. Reading along a line reveals how a single model degrades as one complexity
factor increases. Comparing lines at the same level reveals whether a failure
point is universal or model-specific, a distinction that separates
architectural limits from capacity differences.

Figures~\ref{fig:A1}--\ref{fig:A5} present the five core predictors. Each shows
a steep, monotonic degradation pattern shared across all seven models.
Figures~\ref{fig:A6}--\ref{fig:A9} present the four secondary dimensions, which
are either gated, subsumed, non-monotonic, or proxy-driven. The two groups
together support the sufficiency argument in Section~4.5.

%% ---- A.1 Core Predictors -----------------------------------------------
\subsection*{A.1\quad Core predictors}

The five figures below each isolate one structural bottleneck. In every case the
failure curve is steep, monotonic, and shared across all seven models, confirming
the dimension as a genuine universal predictor rather than a model-specific
artefact.

%% Figure A1 — D4
\begin{figure}[htbp]
  \centering
  \includegraphics[width=\linewidth]{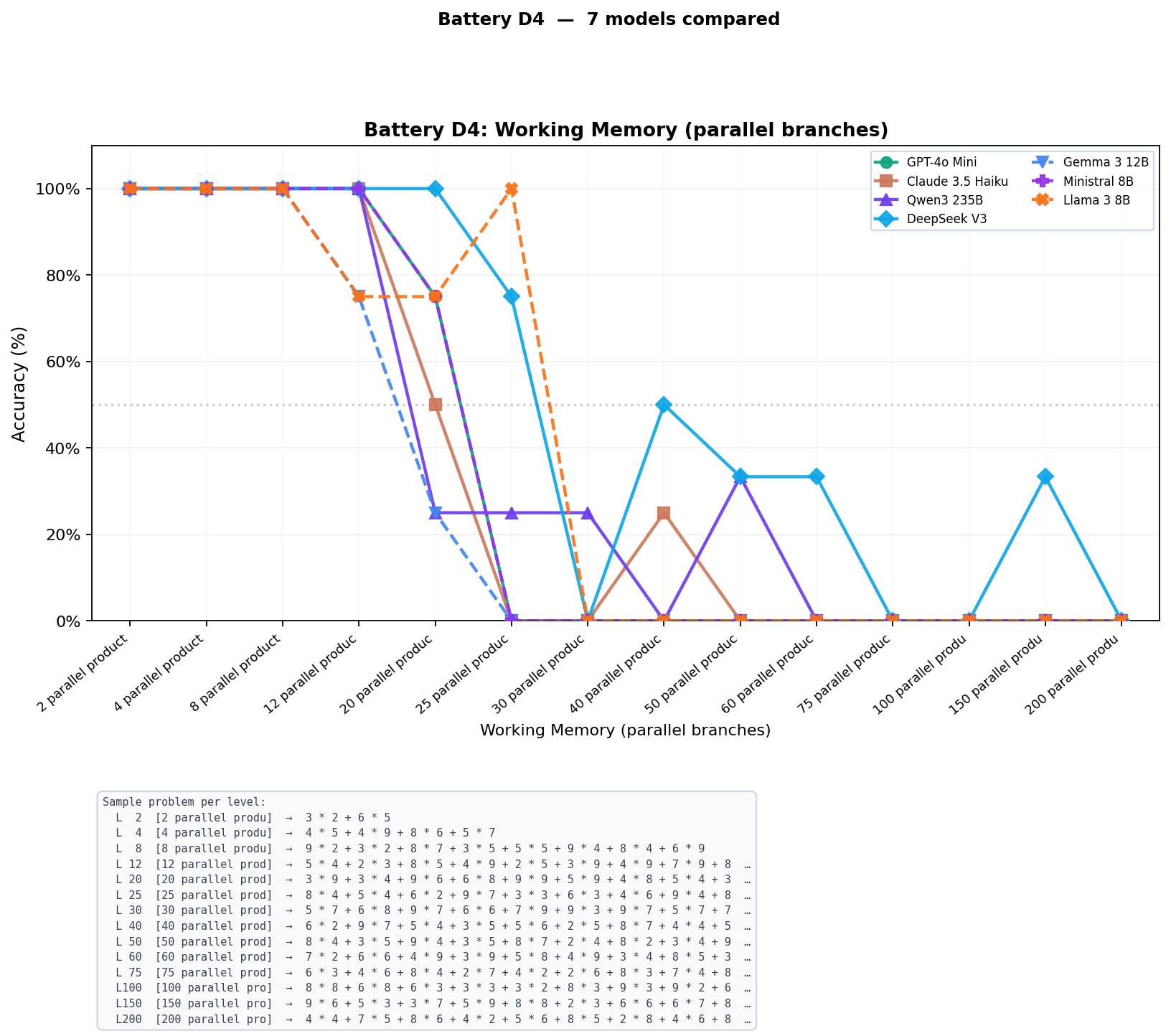}
  \caption{D4, Working Memory (parallel branches: 2\,$\to$\,200).
  Seven models traced across fourteen branch counts: 2, 4, 8, 12, 20, 25, 30,
  40, 50, 60, 75, 100, 150, 200.}
  \label{fig:A1}
\end{figure}

\paragraph{Figure~\ref{fig:A1}, Reading the line graph.}
All seven models form a nearly identical horizontal plateau at 100\% from 2
through 12 branches. Between 15 and 25 branches, every line drops sharply. By 30
branches, every model has reached the 0\% floor and remains there through 200
branches. The collapse band (20--30 branches) is abrupt and scale-invariant:
Qwen3 235B (235B parameters) and Llama 3 8B (8B parameters) fail at identical
thresholds. No model shows any recovery or gradient in the red zone, the
failure is binary and permanent.

\paragraph{Interpretation.}
With $K$ parallel branches, the model must maintain $K$ co-existing intermediate
values simultaneously. Transformer attention has no explicit register mechanism
for this. The scale-invariance, identical thresholds across a 30$\times$
parameter-count difference, is the defining finding of the entire study. This
is not a capacity problem that more parameters solve; it is an architectural
property of how transformers represent and process information. Larger models gain
no additional working memory registers, pointing to a hard structural limit in
self-attention, not a training or capacity constraint.

%% Figure A2 — D2
\begin{figure}[htbp]
  \centering
  \includegraphics[width=\linewidth]{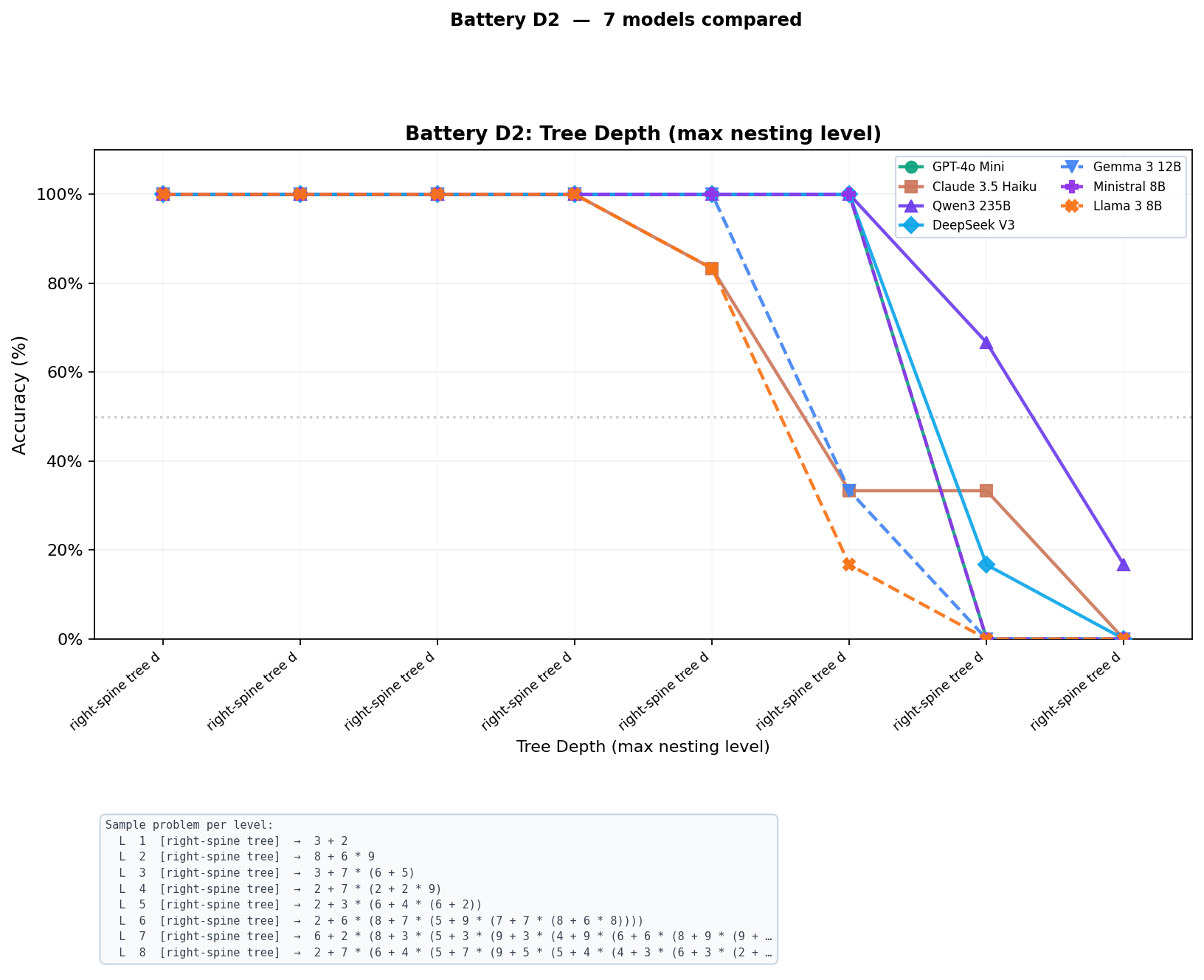}
  \caption{D2, Tree Depth (max nesting level: depth 1\,$\to$\,8).
  Seven models traced across eight nesting depths: 1, 2, 3, 4, 5, 6, 7, 8.}
  \label{fig:A2}
\end{figure}

\paragraph{Figure~\ref{fig:A2}, Reading the line graph.}
All models remain at 100\% accuracy through depth 4. At depth 5, the first
divergence emerges: weaker models (Llama 3 8B, Claude 3.5 Haiku) drop to 67\%
while stronger ones maintain 100\%. The lines compress rightward with each
additional depth level. By depth 6, only GPT-4o Mini, DeepSeek V3, and Qwen3
235B retain non-zero accuracy. Depth 7 shows near-universal collapse except
Qwen3 235B (67\%). Depth 8 produces only one non-zero result: Qwen3 235B at
17\%.

\paragraph{Interpretation.}
At nesting depth $d$, an expression tree requires $2^d$ simultaneous partial
results before the root operation resolves. Depth 5 demands 32 partial results;
depth 6 demands 64; depth 7 demands 128. The rightward-shifting degradation
reflects this exponential demand: stronger models absorb one or two additional
doublings before their working memory ceiling hits. The steep cliff structure ---
not a gradual fade, confirms that models possess discrete working memory limits
rather than capacity degradation. Qwen3 235B's survival at depth 7--8 reflects a
higher effective memory ceiling, not a qualitatively different failure mode.

%% Figure A3 — D8
\begin{figure}[htbp]
  \centering
  \includegraphics[width=\linewidth]{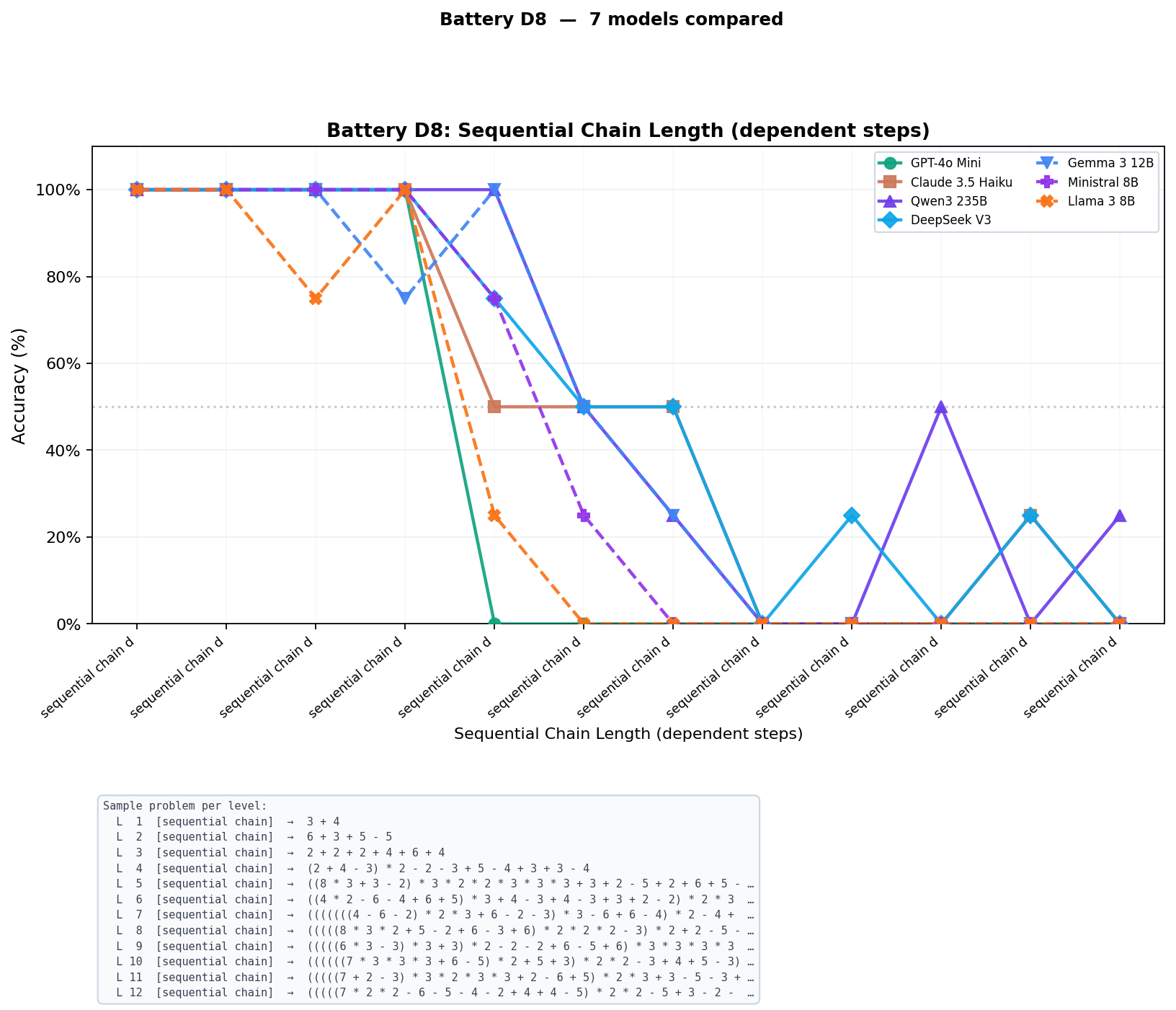}
  \caption{D8, Sequential Chain Length (dependent steps: 1\,$\to$\,12).
  Seven models traced across twelve sequential reasoning levels: steps 1 through
  12.}
  \label{fig:A3}
\end{figure}

\paragraph{Figure~\ref{fig:A3}, Reading the line graph.}
GPT-4o Mini shows the sharpest step-function transition in any suite: 100\%
through step 4, then immediate drop to 0\% from step 5 onward, with no recovery.
This is a hard internal limit. Most other models maintain 100\% through step 4
and begin degrading at step 5--6. Claude 3.5 Haiku and Gemma 3 12B show
scattered recovery (50\% or 25\% at isolated steps) before universal zero by
step 8. Qwen3 235B and DeepSeek V3 show the most resilience, maintaining
non-zero accuracy through step 9, but all models converge to 0\% by step 11--12.
The rightmost zone (steps 10--12) is uniformly zero across all models.

\paragraph{Interpretation.}
Prior two-model studies, capped at 9 steps, classified D8 as only moderately
destructive. The 12-step extension here reveals the complete failure profile: at
sufficient chain length, sequential chaining is as catastrophic as tree depth
(D2). The mechanism is multiplicative error compounding, each dependent step
increases the probability of a cascading error. If step 2 depends on step 1's
output and introduces 5\% error, step 3 inherits both; by step 12, the
accumulated error probability exceeds 50\%. GPT-4o Mini's step-function pattern
(perfect then immediate failure) suggests a hard internal switch for sequential
dependency tracking with no graceful degradation.

%% Figure A4 — D5
\begin{figure}[htbp]
  \centering
  \includegraphics[width=\linewidth]{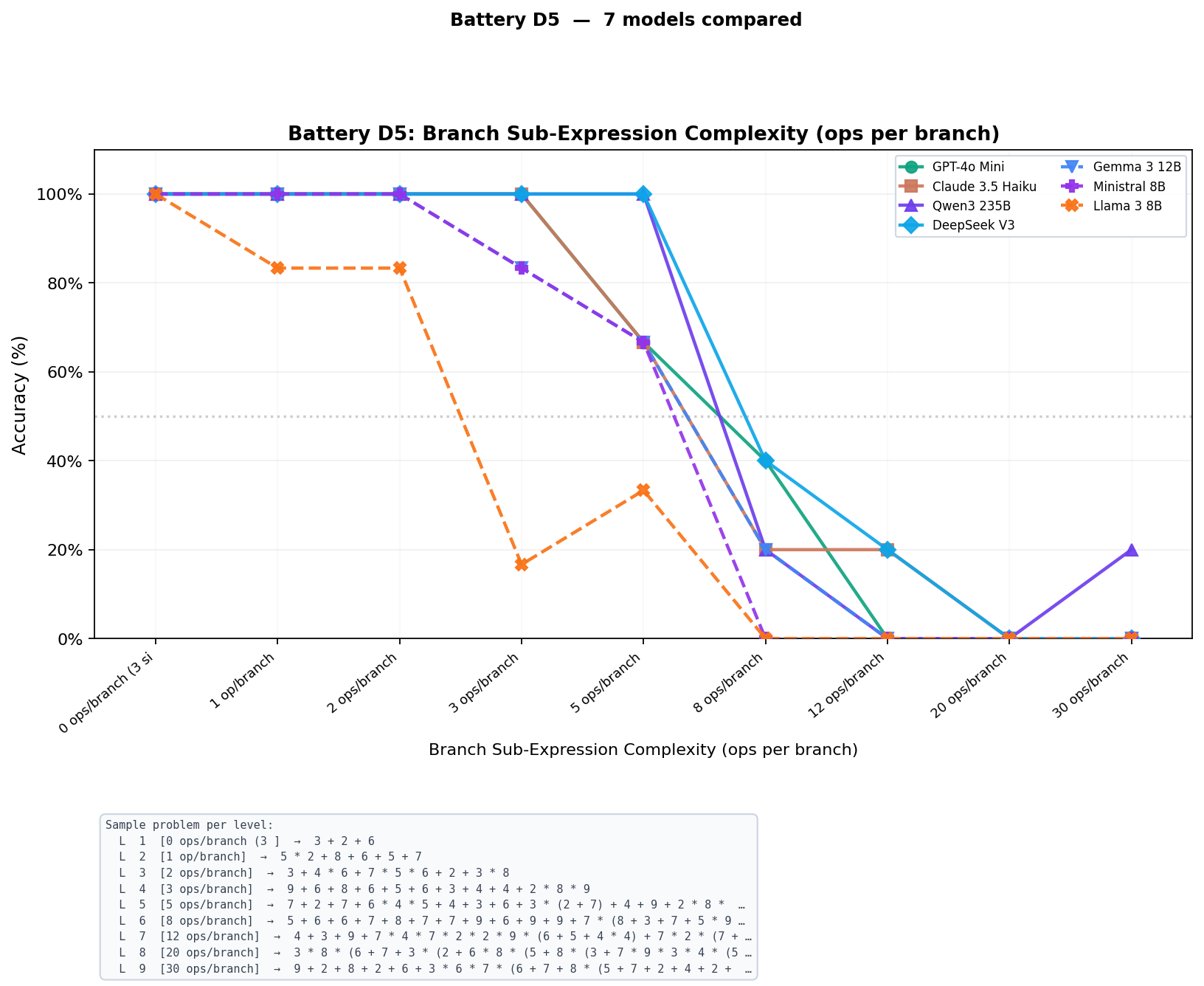}
  \caption{D5, Branch Sub-Expression Complexity (operations per branch:
  0\,$\to$\,30). Seven models traced across nine complexity levels: 0, 1, 2, 3,
  5, 8, 12, 20, 30 operations per branch.}
  \label{fig:A4}
\end{figure}

\paragraph{Figure~\ref{fig:A4}, Reading the line graph.}
All models remain at 100\% through 2~ops/branch. At 3~ops/branch, Llama 3 8B
diverges sharply to 17\%, the earliest model-specific failure in any
dimension, while all others stay at 100\%. At 5~ops/branch, GPT-4o Mini and
Claude 3.5 Haiku drop to 67\% while Qwen3 235B and DeepSeek V3 hold 100\%. The
remaining models collapse in a staggered pattern. By 8~ops/branch, only Qwen3
235B and DeepSeek V3 remain above 20\%. From 12~ops/branch onward, all models
are at or near zero.

\paragraph{Interpretation.}
D5 and D4 both involve branching but capture orthogonal constraints. D4 failure
depends on how many branches coexist; D5 failure depends on how complex each
individual branch becomes before resolving. Llama 3 8B's catastrophic collapse at
3~ops/branch is not mirrored in its D4 profile, the model handles parallel
branch count normally but cannot sustain local reasoning depth within each
branch. This divergence provides the clearest empirical evidence that D4 and D5
measure independent failure modes: a wide shallow tree stresses D4; a narrow deep
tree stresses D5.

%% Figure A5 — D7
\begin{figure}[htbp]
  \centering
  \includegraphics[width=\linewidth]{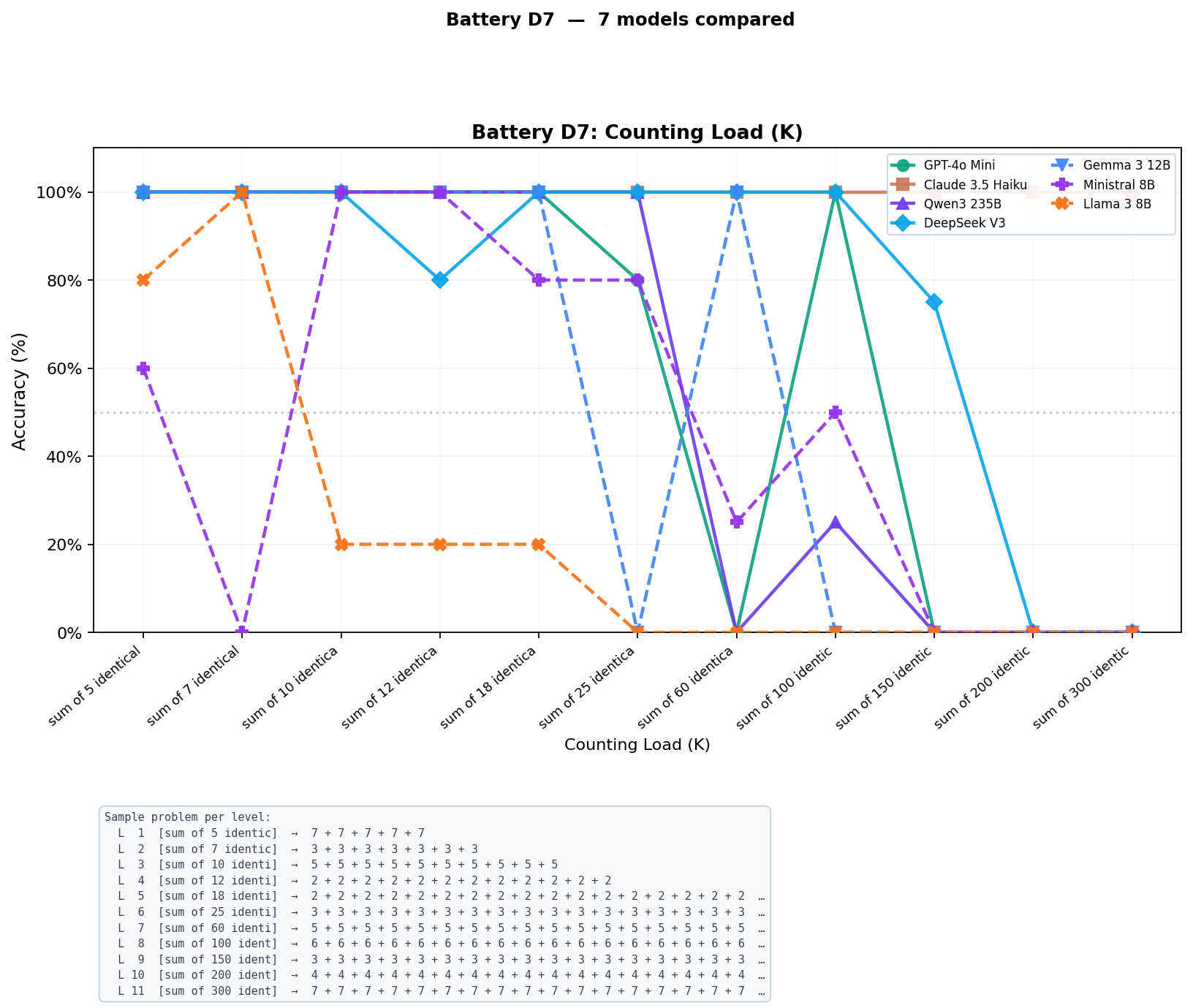}
  \caption{D7, Counting Load ($K$ identical repeated terms:
  $K = 5 \to 300$). Seven models traced across eleven count levels:
  $K = 5, 7, 10, 12, 18, 25, 60, 100, 150, 200, 300$.}
  \label{fig:A5}
\end{figure}

\paragraph{Figure~\ref{fig:A5}, Reading the line graph.}
Claude 3.5 Haiku's line remains at 100\% from $K{=}5$ through $K{=}300$, the
only fully flat line in the entire study. All other models drop dramatically at
$K{=}25$ or $K{=}60$. DeepSeek V3 holds above 50\% through $K{=}60$, declining
thereafter. GPT-4o Mini shows an anomalous spike back to 100\% at $K{=}100$
before returning to zero. Gemma 3 12B, Ministral 8B, and Llama 3 8B are at zero
from $K{=}60$ onward.

\paragraph{Interpretation.}
This dimension shows the widest between-model variance of any suite. Claude 3.5
Haiku's immunity is most plausibly explained by strategy switching: the model
recognises the repeated-addition pattern and converts $K \times \text{value}$,
bypassing the tokenisation ceiling that constrains other models. GPT-4o Mini's
anomalous spike at $K{=}100$ suggests the same bypass fires intermittently rather
than consistently. The paired control suite, identical arithmetic via base-$K$
exponentiation, eliminating any counting demand, confirms that failures in
other models originate in cardinality tracking, not arithmetic computation. This
links to tokenisation limits and positional encoding constraints~\citep{zhang2024tokenization,chang2024}.

\clearpage

%% ---- A.2 Secondary Dimensions ------------------------------------------
\subsection*{A.2\quad Secondary dimensions}

The four dimensions below do not produce the steep, monotonic, universal failure
curves that define core predictors. Each has a distinct reason for its limited
standalone predictive power, documented in the interpretation notes below.

%% Figures A6 + A7 side by side
\begin{figure}[htbp]
  \centering
  \begin{subfigure}[t]{0.48\linewidth}
    \centering
    \includegraphics[width=\linewidth]{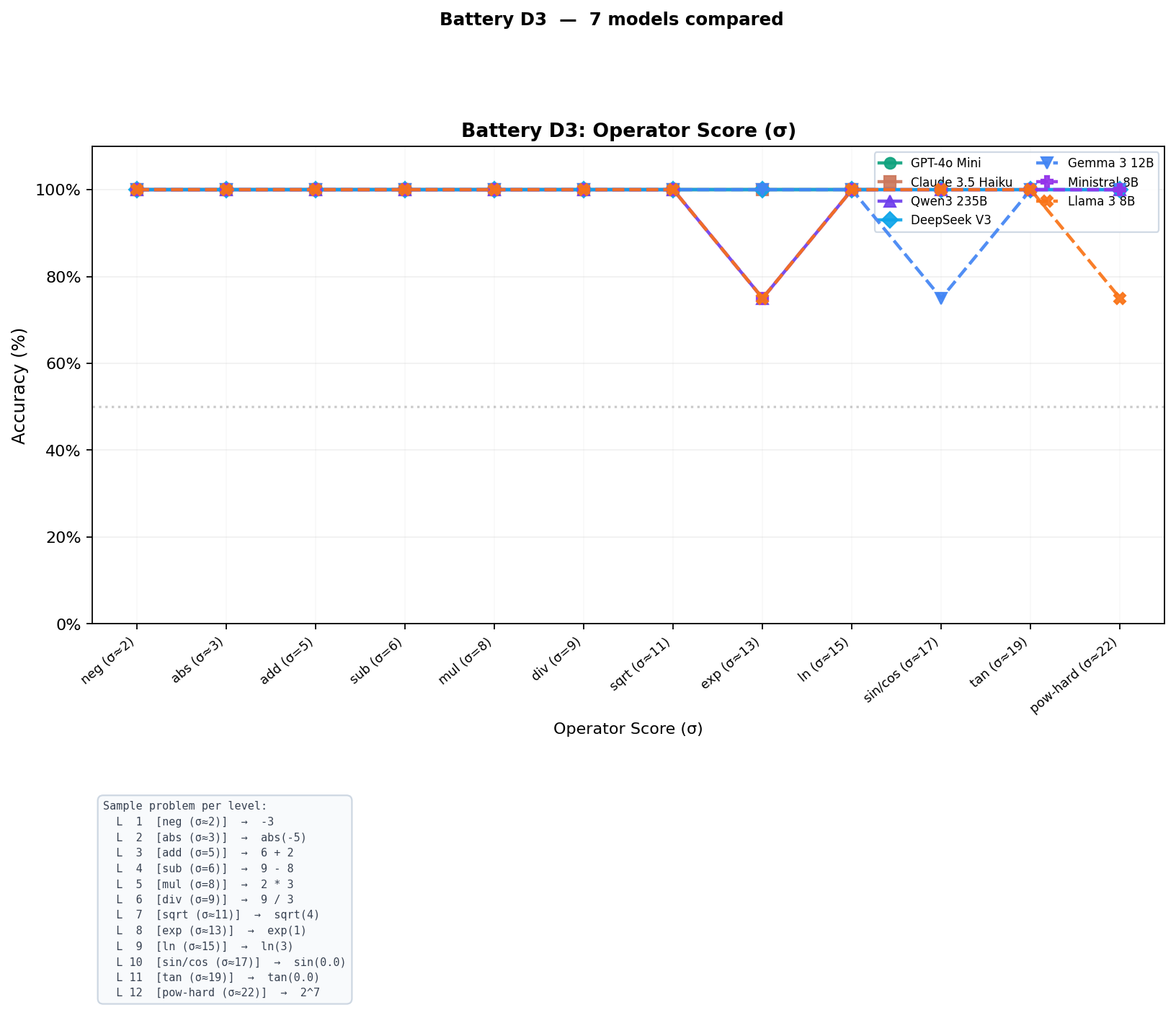}
    \caption{D3, Operator Score $\sigma$
    (\texttt{neg}\,$\to$\,\texttt{pow}).}
    \label{fig:A6}
  \end{subfigure}
  \hfill
  \begin{subfigure}[t]{0.48\linewidth}
    \centering
    \includegraphics[width=\linewidth]{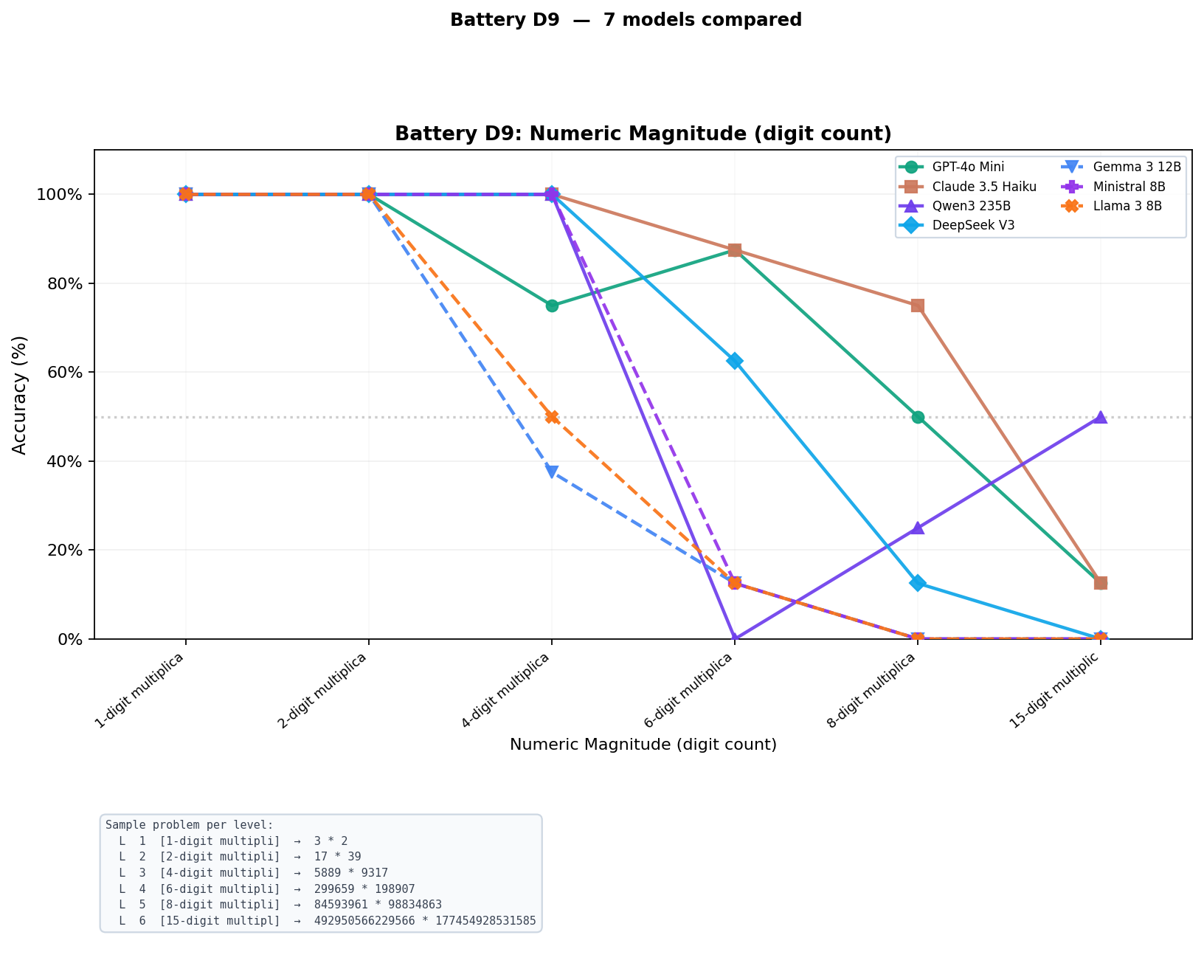}
    \caption{D9, Numeric Magnitude (1\,$\to$\,15 digit multiplication).}
    \label{fig:A7}
  \end{subfigure}
\end{figure}

\paragraph{Figure~\ref{fig:A6}, D3.}
\textbf{Structure:} Seven models traced across twelve operator types ordered by
intrinsic difficulty: \texttt{neg} ($\sigma{=}2$), \texttt{abs} ($\sigma{=}3$),
\texttt{add} ($\sigma{=}5$), \texttt{sub} ($\sigma{=}6$), \texttt{mul/div}
($\sigma{=}8$--9), \texttt{sqrt} ($\sigma{=}11$), \texttt{exp} ($\sigma{=}13$),
\texttt{ln} ($\sigma{=}15$), \texttt{sin/cos} ($\sigma{=}17$), \texttt{tan}
($\sigma{=}19$), \texttt{pow} ($\sigma{=}22$).
\textbf{Reading:} All models maintain 100\% accuracy across the entire operator
range, with only minor dips at \texttt{exp} and \texttt{pow-hard} appearing in
the three weakest models (Gemma 3 12B, Ministral 8B, Llama 3 8B). No model
falls below 75\% even at the suite ceiling ($\sigma{=}22$).
\textbf{Interpretation:} Near-perfect performance across all 12 operators does
not mean operator hardness is unimportant. Rather, it reveals that the
catastrophic failure zone lies above $\sigma{=}22$, in calculus-level territory
requiring differentiation, integration, and Diophantine equations. The small dips
at \texttt{exp} and \texttt{pow-hard} are early signals of the approaching cliff.
The $\sigma$-rank was validated empirically at Spearman $\rho = 0.863$ and is
context-gated: operator failure is catastrophic only when combined with high
complexity in other dimensions.

\paragraph{Figure~\ref{fig:A7}, D9.}
\textbf{Structure:} Seven models traced across six operand sizes: 1-digit,
2-digit, 4-digit, 6-digit, 8-digit, 15-digit multiplication.
\textbf{Reading:} A clean horizontal family-level split. The top two models
(GPT-4o Mini, Claude 3.5 Haiku) remain at 100\% through 8-digit operands, both
dropping to zero at 15 digits. The bottom four models show more varied patterns:
some collapse at 4 digits, others at 6. All models converge to 0\% at 15-digit
multiplication.
\textbf{Interpretation:} D9 does not measure magnitude difficulty in isolation.
Instead, it amplifies D3 via the $O(n)$ vs $O(n^2)$ distinction. With addition,
all models handle large operands well because carry propagation is $O(n)$ ---
linear and learnable. With multiplication, partial products scale $O(n^2)$ ---
each digit pair interacts with all others. By 15 digits, transformer attention
must track $15 \times 15 = 225$ pairwise interactions, far beyond any model's
capacity. D9 is therefore a D3$\times$D9 interaction term, not an independent
predictor~\citep{dziri2024faith}.

%% Figures A8 + A9 side by side
\begin{figure}[htbp]
  \centering
  \begin{subfigure}[t]{0.48\linewidth}
    \centering
    \includegraphics[width=\linewidth]{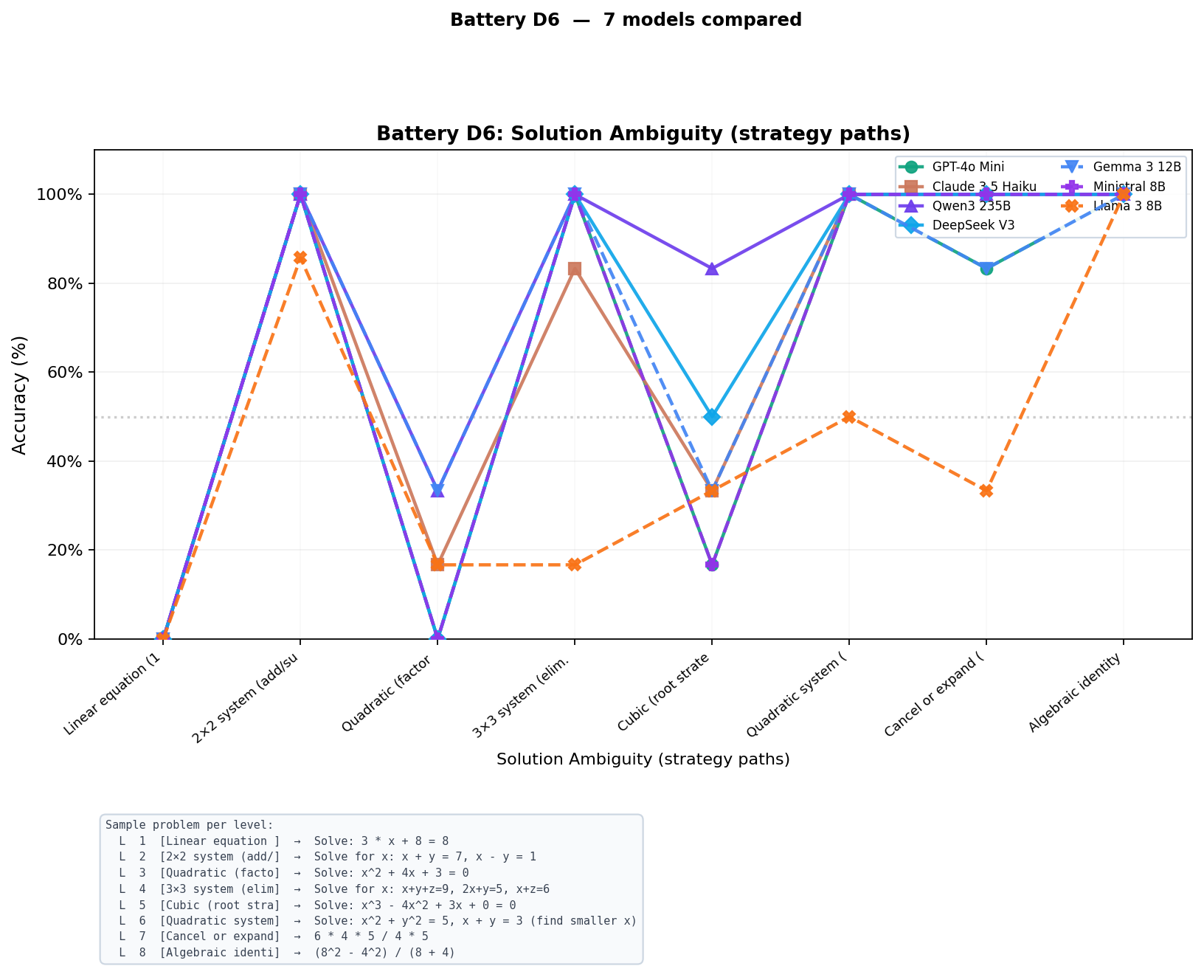}
    \caption{D6, Solution Ambiguity (problem types L1--L8).}
    \label{fig:A8}
  \end{subfigure}
  \hfill
  \begin{subfigure}[t]{0.48\linewidth}
    \centering
    \includegraphics[width=\linewidth]{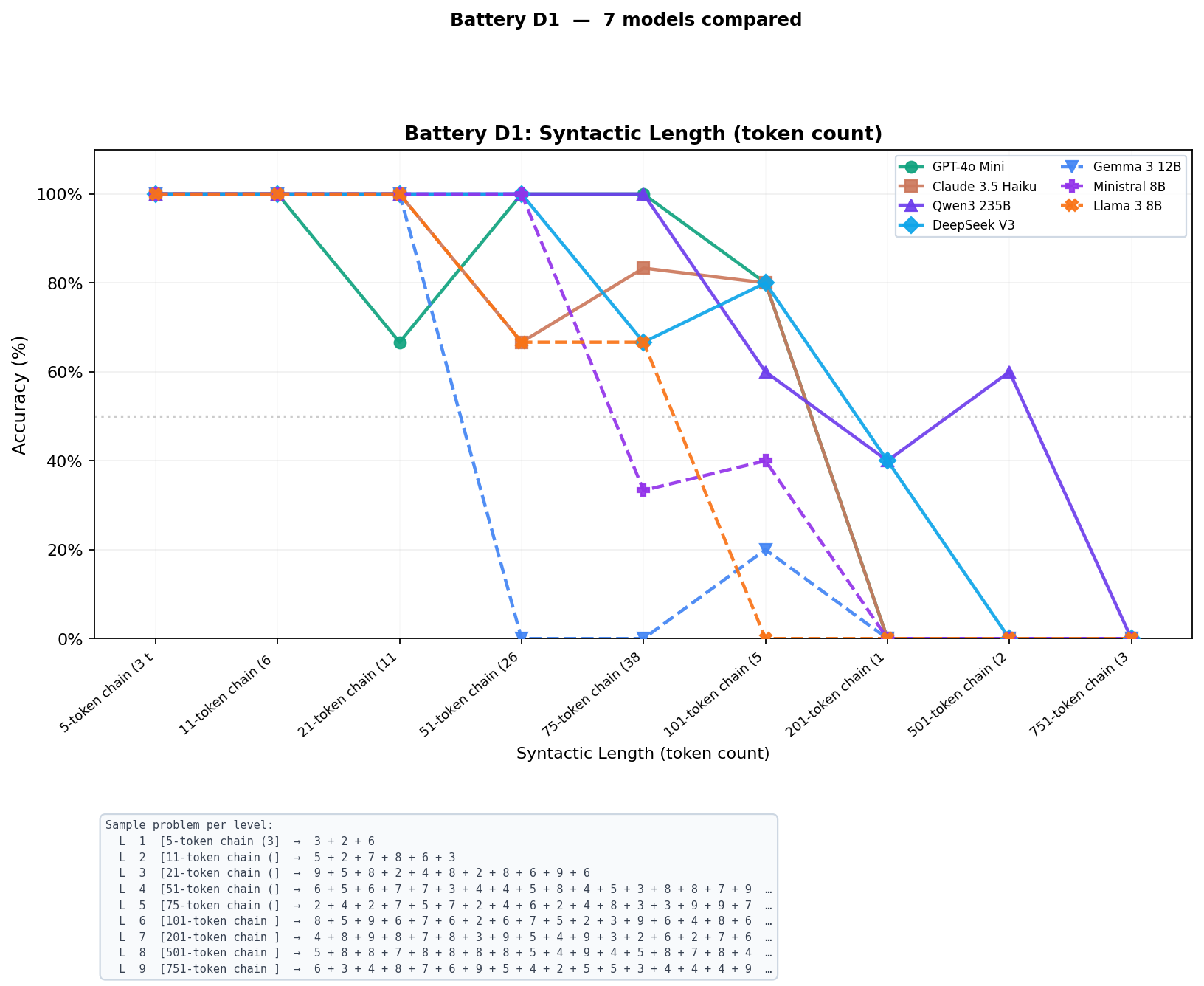}
    \caption{D1, Syntactic Length (5\,$\to$\,751 tokens).}
    \label{fig:A9}
  \end{subfigure}
\end{figure}

\paragraph{Figure~\ref{fig:A8}, D6.}
\textbf{Structure:} Seven models traced across eight algebraic problem types:
linear equation (L1), 2$\times$2 system (L2), quadratic factoring (L3),
3$\times$3 system (L4), cubic (L5), quadratic system (L6), cancel/expand (L7),
algebraic identity (L8).
\textbf{Reading:} The line graph does not produce a smooth monotonic gradient.
Instead, it oscillates: some problem types show high accuracy across all models
while others show widespread failure, with no consistent left-to-right
degradation. Quadratic factoring (L3) is near zero for most models. Algebraic
identities (L8) are uniformly at 100\% across all models. The pattern is
non-monotonic and does not track the nominal strategy-count hypothesis.
\textbf{Interpretation:} Difficulty in D6 does not track solution ambiguity
(number of valid paths) but rather operator familiarity and pattern recognition
--- both already captured by D3. Algebraic identities have many structurally
distinct valid rewrites yet are trivially easy; quadratic factoring has one
standard method yet widely fails. D6 adds no independent predictive signal and
is subsumed by D3~\citep{lample2020deep}.

\paragraph{Figure~\ref{fig:A9}, D1.}
\textbf{Structure:} Seven models traced across nine token count levels: 5, 11,
21, 51, 75, 101, 201, 501, 751 tokens.
\textbf{Reading:} All models maintain accuracy above 67\% through 51 tokens.
From 75 tokens onward, divergence increases sharply. Qwen3 235B shows the most
resilience, holding above 40\% until 501 tokens before dropping to 20\%. GPT-4o
Mini and Claude 3.5 Haiku follow similar degradation curves, both hitting zero
by 201 tokens. Gemma 3 12B, Ministral 8B, and Llama 3 8B collapse earlier, with
Gemma failing completely at 51 tokens. The rightmost segment (501--751 tokens)
shows universal near-zero performance except for Qwen3 235B.
\textbf{Interpretation:} Long expressions strain transformer positional encoding,
introducing new error opportunities at each additional token. However, the
primary driver is not length itself but the nested structures or sequential chains
necessarily implied by expressions exceeding 200 tokens, both independently
captured by D2 and D8. The irregular trajectory in the mid-range (75--201 tokens)
is diagnostic: a genuine causal predictor produces smooth monotonic degradation,
not scattered performance. D1 functions as a downstream proxy, its failures
reflecting D2 and D8 rather than syntactic length as an independent
bottleneck~\citep{markeeva2024clrs}.

\section{Operator difficulty ranking: empirical validation}
\label{app:d3}

 The order of operators according to complexitywas validated on
Qwen-2.5-7B-Instruct before the main dimensions were run. This model was chosen as
the sole validation target because the goal was to confirm the ordering against
measured accuracy, not to compare models. All structural dimensions were held
at their minimum values throughout: depth 1, single-digit operands, no branching,
no sequential chain. With every confound fixed, operator identity is the only
varying factor.

We ran two sample sizes: 20 problems per operator (260 total) and 100 problems per
operator (1,300 total). Both produced Spearman $\rho = 0.863$ and $\rho = 0.868$
respectively against the predicted $\sigma$-rank, with three minor inversions in
eleven consecutive pairs. The ordering was stable across both runs. Hardest
operators, \texttt{pow}, \texttt{tan}, \texttt{sin}, \texttt{cos}, clustered
at the bottom; \texttt{neg}, \texttt{abs}, and \texttt{add} sat firmly at the top,
consistent with predictions from all three derivation signals.

One anomaly appeared in both runs: \texttt{ln} scored 100\% despite its predicted
mid-hard rank ($\sigma{=}15$). The most plausible explanation is that well-formed
\texttt{ln(x)} prompts trigger a lookup-style response, the model retrieves
$\ln(1) = 0$ or $\ln(e) = 1$ as a memorised identity rather than computing the
transcendental. \citet{weber2002} documents the same pattern in human students,
who handle standard logarithm identities correctly while failing on non-routine
instances. This does not invalidate the $\sigma$-rank; it identifies a specific
limitation of single-problem validation rather than an error in the ordering. The
adopted ranking retains \texttt{ln} at $\sigma{=}15$, consistent with the
theoretical derivation. Human error rates on logarithm problems (40--60\%;
\citealp{weber2002}) further support keeping \texttt{ln} in the upper half of the
ordering, as this difficulty propagates into the LLM training signal through
human-written mathematics.

\end{document}